\documentclass[12pt, letterpaper]{article}

\usepackage{amsmath,amsfonts,bm}

\def\eqref#1{equation~\ref{#1}}

\def\1{\bm{1}}

\def\eps{{\epsilon}}

\DeclareMathAlphabet{\mathsfit}{\encodingdefault}{\sfdefault}{m}{sl}
\SetMathAlphabet{\mathsfit}{bold}{\encodingdefault}{\sfdefault}{bx}{n}

\DeclareMathOperator*{\argmax}{arg\,max}

\DeclareMathOperator{\sign}{sign}

\newcommand{\Real}{\mathbb R}

\newcommand{\LinfPGD}{{\text{$\ell_{\infty}$-PGD}}}

\usepackage{hyperref}
\usepackage{url}

\usepackage{amsmath,amsfonts,amssymb,bm}
\usepackage{amsthm}
\usepackage{diagbox}
\usepackage{multirow}
\usepackage{pifont}
\usepackage{makecell}
\usepackage{mathtools}
\usepackage{enumitem}
\usepackage{xspace}

\theoremstyle{definition}

\usepackage{changepage}
\usepackage{graphicx}
\usepackage[linesnumbered,lined,ruled,commentsnumbered]{algorithm2e}

\usepackage[margin=1in]{geometry}
\usepackage[round]{natbib}

\usepackage{authblk}

\newif\ifconsiderlater
\considerlaterfalse

\ifconsiderlater
    \newcommand{\todo}[1]{{\color{cyan} \textbf{XXX [#1] XXX}}}
    \newcommand{\KSComment}[1]{{\color{blue}{[KS: #1]}}}
    \newcommand{\AVComment}[1]{{\color{orange}{[AV: #1]}}}
    \newcommand{\AMComment}[1]{{\color{magenta}{[AM: #1]}}}
\else
    \newcommand{\todo}[1]{}
    \newcommand{\KSComment}[1]{}
    \newcommand{\AVComment}[1]{}
    \newcommand{\AMComment}[1]{}
\fi

\newif\ifSupp
\Supptrue

\newif\ifArxiv
\Arxivtrue

\title{Improving Hierarchical Adversarial Robustness of\\Deep Neural Networks}

\author[1]{Avery Ma\thanks{Work done as an intern at Huawei Noah’s Ark Lab.}}
\author[2]{Aladin Virmaux}
\author[2]{Kevin Scaman}
\author[2]{Juwei Lu}
\affil[1]{University of Toronto, Vector Institute}
\affil[2]{Huawei Noah’s Ark Lab}

\date{}

\begin{document}

\maketitle

\begin{abstract}
Do all adversarial examples have the same consequences?
An autonomous driving system misclassifying a pedestrian as a car may induce a far more dangerous \textemdash and even potentially lethal\textemdash behavior than, for instance, a car as a bus.
In order to better tackle this important problematic, we introduce the concept of \textit{hierarchical} adversarial robustness. 
Given a dataset whose classes can be grouped into \emph{coarse-level} labels, we define hierarchical adversarial examples as the ones leading to a misclassification at the coarse level.
To improve the resistance of neural networks to hierarchical attacks, we introduce a \emph{hierarchical adversarially robust} (HAR) network design that decomposes a single classification task into one coarse and multiple fine classification tasks, before being specifically trained by adversarial defense techniques.
As an alternative to an end-to-end learning approach, we show that HAR significantly improves the robustness of the network against $\ell_2$ and $\ell_{\infty}$ bounded hierarchical attacks on the CIFAR-10 and CIFAR-100 dataset.
\end{abstract}

\section{Introduction}
\label{sec:Introduction}

Deep neural networks (DNNs) are highly vulnerable to attacks based on small modification of the input to the network at test time \citep{szegedy2013intriguing}.
Those adversarial perturbations are carefully crafted in a way that they are imperceptible to human observers, but when added to clean images, can severely degrade the accuracy of the neural network classifier.
Since their discovery, there has been a vast literature proposing various attack and defence techniques for the adversarial settings ~\citep{szegedy2013intriguing, goodfellow2014explaining, kurakin2016adversarial, madry2017towards, wong2020fast}.
These methods constitute important first steps in studying adversarial robustness of neural networks.
However, there exists a fundamental flaw in the way we assess a defence or an attack mechanism.
That is, we overly generalize the mistakes caused by attacks.\AVComment{I am not sure to understand this}\AMComment{does this help?}
Particularly, the current approaches focuses on the scenario where different mistakes caused by the attacks are treated equally.
%
We argue that some context do not allow mistakes to be considered equal.
In CIFAR-100~\citep{krizhevsky2009learning}, it is less problematic to misclassify a pine tree as a oak tree than a fish as a truck.
%


As such, we are motivated to propose the concept of hierarchical adversarial robustness to capture this notion. 
Given a dataset whose classes can be grouped into coarse labels, we define hierarchical adversarial examples as the ones leading to a misclassification at the coarse level; and we present a variant of the projected gradient descent (PGD) adversaries~\citep{madry2017towards} to find hierarchical adversarial examples. 
Finally, we introduce a simple and principled hierarchical adversarially robust (HAR) network which decomposes the end-to-end  learning task into one coarse and multiple fine classification tasks, before being trained by adversarial defence techniques.
%
%
Our contributions are
\begin{itemize}

 	\item We introduce the concept of hierarchical adversarial examples: a special case of the standard adversarial examples which causes mistakes at the coarse level (Section~\ref{sec:Hierarchical_Adversarial_Examples}).
 	
 	\item Motivated by the empirical observation that untargeted PGD attacks yield misclassifications within the same coarse label (Section~\ref{sec:observations_untargeted_pgd}), we present a \textit{worst-case} targeted PGD attack to find hierarchical adversarial examples. The attack iterates through all candidate fine labels until a successful misclassification into the desired target (Section~\ref{sec:generating_hierarchical_perturbation}).
 	
    \item We propose a novel architectural approach, HAR network, for improving the hierarchical adversarial robustness of deep neural networks (Section~\ref{sec:Method}). 
    We empirically show that HAR networks significantly improve the hierarchical adversarial robustness against $\ell_\infty$ attacks ($\eps = \frac{8}{255}$) and $\ell_2$ attacks ($\eps = 0.5$) on CIFAR-10 and CIFAR-100 (Section~\ref{sec:Experiments} and \ifSupp  Appendices~\ref{sec:appendix_cifar100_l2}, \ref{sec:appendix_cifar10}\else the supplementary material \fi).
    
    \item We benchmark using untargeted PGD attacks as well as the proposed worst-case targeted PGD attack. 
    In particular, we include an extensive empirical study on the improved hierarchical robustness of HAR by evaluating against attacks with different $\ell_p$ norms, varying PGD iterations and $\eps$. 
    %
    The result shows that the proposed worst-case targeted attack provides a more accurate empirical representation of the hierarchical adversarial robustness of the model (Section~\ref{sec:exp_untargeted_targeted_attack}).
    
    \item We show that the iterative targeted attack formulated based on the coarse network are weaker hierarchical adversarial examples compared to the ones generated using the entire HAR network (Section~\ref{sec:exp_targeted_coarse}).

\end{itemize}

\section{Hierarchical Adversarial Examples}
\label{sec:Hierarchical_Adversarial_Examples}
\AVComment{in my opinion this beginning is not great. Maybe restructure the logic path in
1) advancement in DNN make network more complex and more vulnerable (?) 2) more data with more and more classes 3) need to have some hierarchy within these classes}
The advancement in DNN image classifiers is accompanied by the increasing complexity of the network design \citep{szegedy2016inception, he2016deep}, and those intricate networks has provided state-of-the-art results on many benchmark tasks~\citep{deng2009imagenet, geiger2013vision, cordts2016cityscapes, everingham2015pascal}. 
Unfortunately, the discovery of adversarial examples has revealed that neural networks are extremely vulnerable to maliciously perturbed inputs at test time~\citep{szegedy2013intriguing}.
This makes it difficult to apply DNN-based techniques in mission-critical and safety-critical areas.
%

Another important development along with the advancement of DNN is the growing complexity of the dataset, both in size and in number of classes: i.e. from the 10-class MNIST dataset to the 1000-class ImageNet dataset. 
\AVComment{growing complexity of datasets both in size and in number of classes}%
\AMComment{edited}%
As the complexity of the dataset increases exponentially, dataset can often be divided into several coarse classes where each coarse class consists of multiple fine classes. 
In this paper, we use the term label and class interchangeably.

The concept of which an input image is first classified into coarse labels and then into fine labels are referred to as \emph{hierarchical classification}~\citep{tousch2012semantic}. 
Intuitively, the visual separability between groups of fine labels can be highly uneven within a given dataset, and thus some coarse labels are more difficult to distinguish than others. 
This motivates the use of more dedicated classifiers for specific groups of classes, allowing the coarse labels to provide information on similarities between the fine labels at an intermediate stage.
The class hierarchy can be formed in different ways, and it can be learned strategically for optimal performance of the downstream task~\citep{deng2011fast}.
Note that it is also a valid strategy to create a customized class hierarchy and thus be able to deal with sensitive missclassification.
To illustrate our work, we use the predefined class hierarchy of the CIFAR-10 and the CIFAR-100 dataset~\citep{krizhevsky2009learning}: fine labels are grouped into coarse labels by semantic similarities. 
%

%


All prior work on adversarial examples for neural networks, regardless of defences or attacks, focuses on the scenario where all misclassifications are considered equally~\citep{szegedy2013intriguing, goodfellow2014explaining, kurakin2016adversarial, madry2017towards, wong2020fast}. \AVComment{citations needed}\AMComment{added}
However, in practice, this notion overly generalizes the damage caused by different types of attacks.
For example, in an autonomous driving system, confusing a perturbed image of a traffic sign as a pedestrian should not be treated the same way as confusing a bus as a pickup truck. 
The former raises a major security threat for practical machine learning applications, whereas the latter has very little impact to the underlying task.
Moreover, misclassification across different coarse labels poses potential ethical concerns when the dataset involves sensitive features such as different ethnicities, genders, people with disabilities and age groups.
\ifArxiv
    \begin{adjustwidth}{1cm}{1cm}
    \textit{Mistakes across coarse classes lead to much more severe consequences compared to mistakes within coarse classes.}
    \end{adjustwidth}
\else 
    \begin{adjustwidth}{0.1cm}{0.1cm}
    \textit{Mistakes across coarse classes lead to much more severe consequences compared to mistakes within coarse classes.}
    \end{adjustwidth}
\fi
As such, to capture this different notion of attacks, we propose the term \textit{hierarchical} adversarial examples. They are a specific case of adversarial examples where the resulting misclassification occurs between fine labels that come from different coarse labels. 

Here, we provide a clear definition of the hierarchical adversarial examples to differentiate it from the standard adversarial examples. We begin with the notation for the classifier.
%
%
Consider a neural network $F(x): \Real^d \rightarrow \Real^c$ with a softmax as its last layer~\citep{HastieTibshiraniFriedman2009}, 
where $d$ and $c$ denote the input dimension and the number of classes, respectively. 
The prediction is given by $\argmax_i F(x)_i$.
%

In the hierarchical classification framework, the classes are categorized (e.g. by the user) into fine classes and coarse classes \footnote{We could go beyond this 2-level hierarchy. Here we keep the presentation simple for didactic purposes.}.
The dataset consists of image and fine label pairs: $\left \{ x, y \right \}_n$. 
In the later, we use the set theory symbol $\in$ to characterize the relationship between a fine and a coarse label: $y \in z$ if the fine label $y$ is part of the coarse class $z$.
%
%
Note that this relation holds for both disjoint and overlapping coarse classes.
%
%
Given an input data $x$, suppose its true coarse and fine labels are $z^*$ and $y^*$ respectively.
Under the setting defined above, a hierarchical adversarial example must satisfy all the following properties:

\ifArxiv
    \begin{itemize}[leftmargin=7mm]
        \item the unperturbed data $x$ is correctly classified by the classifier: $\argmax_i F(x)_i = y^*$;
        \item the perturbed data $x' = x + \delta$ is perceptually indistinguishable from the original data $x$;
        \item the perturbed data $x'$ is classified incorrectly: $\argmax_i F(x')_i = y'$ where $y' \neq y^*$;
        \item the misclassified label belongs to a different coarse class: $y' \not\in z^*$.
    \end{itemize}
\else 
    \begin{itemize}[itemsep=0mm, , topsep=0mm]
        \item the unperturbed input data $x$ is correctly classified by the classifier: $\argmax_i F(x)_i = y^*$;
        \item the perturbed data $x' = x + \delta$ is perceptually indistinguishable from the original data $x$;
        \item the perturbed data $x'$ is classified incorrectly: $\argmax_i F(x')_i = y'$ where $y' \neq y^*$;
        \item the misclassified label belongs to a different coarse class: $y' \not\in z^*$.
    \end{itemize}
\fi
Notice that satisfying the first three properties is sufficient to define a standard adversarial examples, and that hierarchical adversarial examples are special cases of adversarial examples.
It is worth mentioning that measuring perceptual distance can be difficult \citep{li2003discovery}, thus the second property is often replaced by limiting that the adversary can only modify any input $x$ to $x + \delta$ with $\delta \in \Delta$.
Commonly used constraints are $\eps$-balls w.r.t. $\ell_p$-norms, though other constraint sets have been used too \citep{wong2019wasserstein}.
In this work, we focus on $\ell_\infty$- and $\ell_2$-norm attacks.



\subsection{Empirical Observations on Untargeted Attacks}
\label{sec:observations_untargeted_pgd}
A common class of attack techniques are gradient-based attacks, such as FGSM~\citep{goodfellow2014explaining}, BIM~\citep{kurakin2016adversarial} and PGD~\citep{madry2017towards}, that utilize gradient (first-order) information of the network to compute perturbations. 
%
Such methods are motivated by linearizing the loss function and solving for the perturbation that optimizes the loss subject to the $\ell_p$-norm constraint. 
Their popularity is largely due to its simplicity, because the optimization objective can be accomplished in closed form at the cost of one back-propagation.
In the context of our work, an important notion is that the attack can be either targeted or untargeted. 
The untargeted attack aims to perturb the image in a way that the altered image will be classified incorrectly.
On the other hand, the goal of the targeted attack is to cause misclassification into a specified target class.

Despite a plethora of attack techniques has been proposed, PGD remains as a fundamental algorithm for many intricate attacks and a standard empirical approach for measuring model robustness \citep{gowal2019alternative, mao2020composite, tashiro2020diversity, croce2020reliable}. 
For instance, the recent SOTA method AutoAttack \citep{croce2020reliable} consists an ensemble of powerful attacks where PGD is directly used for evaluating model's robustness against untargeted adversaries.

To motivate the hierarchical adversarial perturbations in Section \ref{sec:generating_hierarchical_perturbation}, we start with an empirical analysis on the mistakes made by networks on clean and perturbed images.
We found that untargeted $\ell_\infty$-norm PGD20 attacks ($\eps = \frac{8}{255}$) are more likely to yield misclassifications within the same coarse label, captured by the results shown in Table \ref{table:wrong_child_but_correct_parent} where we compute the percentage of misclassified fine labels that are still correctly classified at the coarse level: $\frac{\# \: \text{still classified correctly at the coarse level}}{\# \: \text{misclassified at the fine level}} \times 100\%$.
The two robustified models are trained using PGD10 adversarial training (ADV) \cite{madry2017towards} and TRADES \cite{zhang2019theoretically}.
We make two important observations from the results in Table \ref{table:wrong_child_but_correct_parent}.
%
%
First, note that misclassifications from the untargeted attack are more likely to be within the same coarse label.
This is compared to the chance of being classified as a random coarse label, which is $50\%$ and $5\%$ for CIFAR-10 and CIFAR-100 respectively.
%
%
Second, notice that given an image that is misclassified at the fine level, whether perturbed or not does not significantly change its likelihood of being misclassified at the coarse level.
For example, we observe that for a TRADES-robustified model trained on CIFAR-10, around $77\%$ of both misclassified clean images and PGD-perturbed images are still correctly classified at the coarse level.
%
%
%
This suggests that in terms of degrading hierarchical robustness, an untargeted PGD-perturbed input is just as ineffective as an unperturbed one.
This motivates us to adopt a new approach for generating hierarchical adversarial perturbations, as well as a metric for robustness evaluation.

\begin{table}[t]
\begin{footnotesize}
\caption{Percentage of the misclassified fine labels that are still correctly classified at the coarse level, e.g., misclassifying a ship as an airplane which is also in the``vehicle'' coarse label of CIFAR-10.}
\label{table:wrong_child_but_correct_parent}
\vskip 0.15in
\begin{center}
\setlength{\tabcolsep}{8pt} 
\renewcommand{\arraystretch}{1.3} 
\begin{tabular}{llllll}
\multicolumn{1}{c}{\multirow{2}{*}{Method}} 	   & \multicolumn{2}{c}{CIFAR-10} 	    & \multicolumn{2}{c}{CIFAR-100}  \\ \cline{2-5}
          & Clean    & PGD20    & Clean  & PGD20 \\ \Xhline{2\arrayrulewidth}
Standard  & $84.46\%$  & $80.00\%$  & $33.71\%$ & $36.80\%$  \\
ADV       & $83.75\%$  & $75.56\%$  & $35.57\%$ & $39.06\%$  \\
TRADES    & $77.39\%$  & $77.14\%$  & $34.84\%$ & $38.22\%$  \\\Xhline{2\arrayrulewidth}
\end{tabular}
\end{center}
\end{footnotesize}
\end{table}

\subsection{Generating Hierarchical Adversarial Perturbations}
\label{sec:generating_hierarchical_perturbation}

%
The main idea of gradient-based attacks can be summarized as follows. Given the prediction of $F(x)$ and a target label $y$, the loss function of the model is denoted by $\ell(x, y) \triangleq \ell(F(x), y)$, e.g., a cross-entropy loss. Here, we omit the network parameter $w$ in the loss because it is assumed to be fixed while generating adversarial perturbations.
%
Note that the choice of $y$ and whether to maximize or minimize the loss depend on if the attack is targeted or untargeted.
For a targeted $\ell_\infty$ attack, gradient-based methods rely on the opposite direction of the loss gradient, $-\sign{\nabla_x\ell(x,y)}$, to solve for the perturbation that \textbf{minimizes} the loss with respect to a non-true target label ($y \neq y^*$).
%
%
Despite its simplicity, gradient-based attacks are highly effective at finding $\ell_p$-bounded perturbations that lead to misclassifications. 

In our work, we introduce a simple variant of the PGD adversary to find hierarchical adversarial examples. 
Given an input image with true coarse and fine labels $z^*$ and $y^*$ respectively. Let $x_j$ denote the perturbed input at iteration $j$, we define:
\begin{equation}
    x_{j+1} = \Pi_{B_{\infty}\left(x, \eps\right)}\left\{x_j - \alpha \sign{(\nabla_{x} \ell(x_j, \hat{y}))}\right\}
    \label{eq:targeted_attack}
\end{equation}
where the target label $\hat{y}$ comes from a different coarse class: $\hat{y} \not\in z^*$.
Algorithm \ref{alg:hierarchical_example} summarize the procedures for generating an $\ell_\infty$-constrained hierarchical adversarial examples.
The projection operator $\Pi$ after each iteration ensures that the perturbation is in an $\eps$-neighbourhood of the original image. 
We also adopt the random initialization in PGD attacks \citep{madry2017towards}: $x_{0} = x + \eta$, where $ \eta = (\eta_1, \eta_2, \dotsc, \eta_d)^\top$ and $\eta_i \sim \mathcal{U}(-\eps, \eps)$. 
%

\begin{algorithm}
\SetKwData{Left}{left}\SetKwData{This}{this}\SetKwData{Up}{up}
\SetKwInOut{Input}{Input}\SetKwInOut{Output}{Output}
\textbf{Input:} A pair of input data $(x, y^*)$, where fine label $y^*$ belongs to the coarse label $z^*$; a neural network $F(\cdot)$; loss function $\ell(\cdot)$; $\ell_\infty$ constraint of $\eps$; number of PGD iterations $k$; PGD step-size $\alpha$.
\BlankLine
\textbf{Define:}  $S = \{y \mid y \not\in z^* \}$, a collection of all fine labels that do not belong in the coarse label $z^*$\;
\For{$\hat{y} \in S$}{
    $x_{0} \leftarrow x + \eta$, where $ \eta \leftarrow (\eta_1, \eta_2, \dotsc, \eta_d)^\top$ 
        and $\eta_i \sim \mathcal{U}(-\eps, \eps)$.\\ 
    \For{$j = 0,\dotsc,k-1$ }{

        $x_{j+1} = \Pi_{B_\infty\left(x, \eps\right)}\left\{x_{j} - \alpha \sign{(\nabla_{x} \ell(x_j, \hat{y}))}\right\}$ 
        where $\Pi$ is the projection operator.\\
    }
    \eIf{$\argmax_i F(x_k)_i = \hat{y}$}{
      Terminate (successful attack)\;
      }{
      $S \setminus \hat{y}$\;
      \If{$S$ is empty}{
      Terminate (failed attack)\;
      }}
    
}
\caption{A \textit{worst-case} approach for generating $\ell_\infty$-bounded hierarchical adversarial example based on a targeted PGD attack.}
\label{alg:hierarchical_example}
\end{algorithm}

There are several approaches to choose the target class~\citep{carlini2017towards}. The target class can be chosen in an \textit{average-case} approach where the class is selected uniformly at random among all eligible labels. Alternatively, they can be chosen in a strategic way, a \textit{best-case} attack, to find the target class which requires the least number of PGD iterations for misclassifications. In our work, we consider a \textit{worst-case} attack by iterating through all candidate target labels, i.e., fine labels that do not belong in the same coarse class. This iterative targeted attack process terminates under two conditions: 1. perturbation results in a successful targeted misclassification; 2. all candidate fine labels have been used as targets.

\subsection{Related Work on Hierarchical Classification}

In image classification domain, there is a sizable body of work exploiting class hierarchy of the dataset~\citep{tousch2012semantic}. 
For classification with a large number of classes, it is a common technique to divide the end-to-end learning task into multiple classifications based on the semantic hierarchy of the labels~\citep{marszalek2008constructing, liu2013probabilistic, deng2012hedging}.
They are motivated by the intuition that some coarse labels are more difficult to distinguish than others, and specific category of classes requires more dedicated classifiers.
A popular hierarchy structure is to divide the fine labels into a label tree with root nodes and leaf nodes, and \citet{deng2011fast} propose an efficient technique to simultaneously determine the structure of the tree as well as learning the classifier for each node in the tree.
It is also common to use the predefined hierarchy of the dataset~\citep{deng2012hedging}.
%

%

\section{Hierarchical Adversarially Robust (HAR) Network}
\label{sec:Method}
To improve the hierarchical adversarial robustness of neural networks, we propose a simple and principled hierarchical adversarially robust (HAR) network which decompose the end-to-end robust learning task into two parts. 
First, we initialize a neural network for the coarse classification task, along with multiple networks for the 
fine classification tasks. 
Next, all the networks are trained using adversarial defence techniques to improve the robustness of their task at hand. 
The final probability distribution of all the fine classes are computed based on Bayes Theorem. 
For brevity, we use coarse neural network (CNN) and fine neural network (FNN) to denote the two different types of networks.
Intuitively, the HAR network design benefits from a single robustified CNN with improved robustness between coarse classes, and multiple robustified FNN with improved the robustness between visually similar fine classes.
\begin{figure*}[t!]
\begin{center}
    \includegraphics[clip, trim=3cm 21cm 3cm 1cm, width=1.00\textwidth]{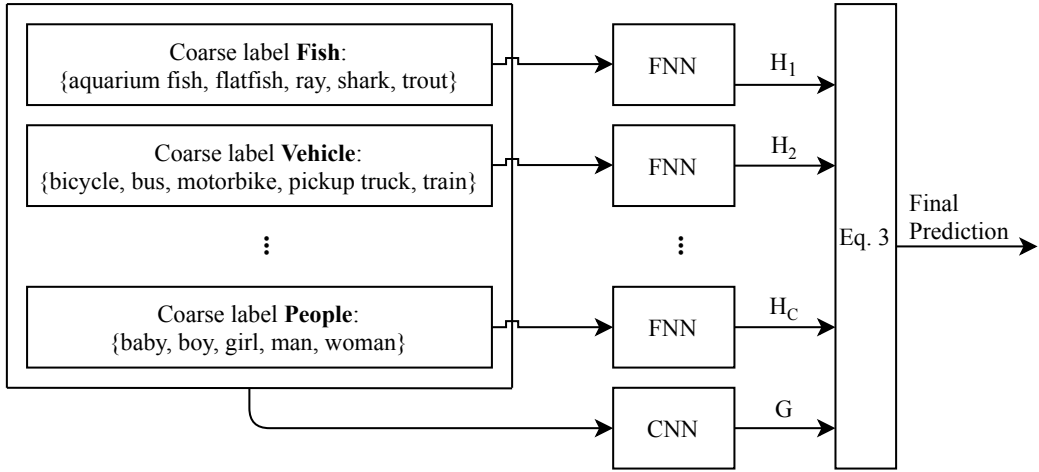}
\caption{Pipeline of the proposed HAR network design to improve hierarchical adversarial robustness of the neural network. The final prediction of the HAR network is computed using Equation \ref{eq:prediction} based on the CNN's prediction of the coarse classes ($G(x) = \left [ g_1, \dotsc, g_{\text{c}} \right ]$) and the FNNs’ prediction of the fine classes ($H_{i}(x), \dotsc, H_{c}(x)$).}
\end{center}
\end{figure*}
\subsection{Architecture Design of HAR}
Instead of the traditional flat design of the neural network, HAR consists of one CNN for the coarse labels and several FNNs for the fine labels. 
Note that there is a one-to-one correspondence between a particular FNN and a specific group of fine labels.
Such a module design mimics the hierarchical structure of the dataset where the fine classes are grouped into coarse classes.
Recall that our definition of the neural network $F(x)$ includes the softmax function as its last layer, so the output of the network can be interpreted as the probability distribution of the classes: $P(y \mid x)$. Conditioned on the probability of the coarse class, we can define the fine class probability as 
\begin{equation}
    P(y \mid x) =  P(y \mid x,z) P(z \mid x).
\end{equation}





Here, the probability distribution of the fine classes are computed as the product of two terms. 
Given an input $x$, the first term $P(y \mid x,z)$ represents the probability of $x$ being a fine label $y$ where $y$ belongs to a coarse class $z$. 
This is essentially FNNs' prediction of the fine classes within a coarse category.
The second term $P(z \mid x)$ represents the probability of $x$ being a coarse label $z$, and it can be understood as the prediction from the CNN. With this decomposition of the original learning task, we can reconstruct the fine label distribution by probabilistically combining the predictions from the different networks.
%


An important advantage of this flexible modular network design is that it allows us to train the component networks using adversarial defence techniques to improve the robustness of their associated task.
Especially, a robustified coarse neural network leads to improved hierarchical adversarial robustness between coarse labels.
During training, each component of the HAR network is trained independently, allowing them to be trained in parallel. We use the entire dataset with the coarse labels, $\left \{x, z \right\}$, to train the coarse class network $G(x)$, followed by training multiple fine class network $H(x)$ using only a portion of the dataset.

The inference procedure can be described as follows.
Suppose the number of coarse classes in a dataset is $C$, and each coarse class contains $j$ number of fine classes. 
Similar to the definition of $F(x)$, we use $G(x)$ to denote the output of the CNN: $G(x) = \left [ g_1, \dotsc, g_{\text{c}} \right ]$.
We use $H_{i}(x)$ to denote the output of the FNN: $H_{i}(x) = \left [ h_{1}^{i}, \dotsc, h_{j}^{i} \right ]$, where $j$ is an positive integer indicating the number of fine classes in the coarse class $i$.
%
%
In this setting, the output of the combined neural network is: 
\begin{equation}
  F(x) = \left [ g_1H_{1}(x), \dotsc, g_\text{C}H_\text{C}(x) \right ].\label{eq:prediction}
\end{equation}
\subsection{Related Work on Adversarial Defence Methods}
%
A large body of defence mechanisms have been proposed for the adversarial setting.
Adversarial training~\citep{szegedy2013intriguing} is one of the standard approaches for improving the robustness of deep neural networks against adversarial examples.
It is a data augmentation method that replaces unperturbed training data with adversarial examples and updates the network with the replaced data points.
Intuitively, this procedure encourages the DNN not to make the same mistakes against an adversary.
By adding sufficiently enough adversarial examples, the network gradually becomes robust to the attack it was trained on.
Existing adversarial training methods ~\citep{szegedy2013intriguing, goodfellow2014explaining, kurakin2016adversarial, madry2017towards, wong2020fast} differ in the adversaries used in the training.
Another related line of adversarial defence methods focuses on regularizing the loss function instead of data augmentation.
TRADES~\citep{zhang2019theoretically} introduces a regularization term that penalizes the difference between the output of the model on a training data and its corresponding adversarial example. 
The regularized loss consists of a standard cross-entropy loss on the unperturbed data and a KL-divergence term measuring the difference between the distribution of clean training data and adversarially perturbed training data.
\section{Experiments}
\label{sec:Experiments}
In this section, we evaluate the hierarchical adversarial robustness of the HAR network design, incorporating two popular adversarial defence methods: adversarial training with PGD10 adversaries \citep{madry2017towards} and TRADES \citep{zhang2019theoretically}. 
In this section, we focus on evaluations based on $\ell_\infty$ norm attacks on the CIFAR-100 dataset, and defer evaluations on $\ell_2$ of CIFAR-100 in \ifSupp Appendix \ref{sec:appendix_cifar100_l2}\else the supplementary material \fi.
Additionally, we include evaluations on the small-sized CIFAR-10 dataset and the medium-sized CIFAR-100-5x5 dataset (a 25-class subset of CIFAR-100) in \ifSupp Appendix \ref{sec:appendix_cifar10}\else the supplementary material \fi.
The result shows that the standard untargeted PGD attack is insufficient at empirically evaluating model's hierarchical adversarial robustness, as shown by the significant drop in coarse class accuracy from the proposed worst-case targeted attack.
Compared to the traditional flat design of neural network, our experiments show that HAR leads to a significant improvement in hierarchical adversarial robustness under various targeted and untargeted attacks (Section \ref{sec:exp_untargeted_targeted_attack}).
Lastly, we show that one can generate hierarchical adversarial examples based on the CNN part of the HAR network under the white-box threat model, but they are not as strong as the proposed worst-case targeted attack in degrading model's hierarchical robustness (Section \ref{sec:exp_targeted_coarse}).

\subsection{Evaluation Setup}
We use network architectures from the ResNet family \citep{he2016deep} on the CIFAR-100 dataset. The hierarchical structure of classes within the dataset is illustrated in \ifSupp Table~\ref{table:cifar10_cifar100}\else the supplementary material \fi. 
%
To establish a baseline, we train ResNet50 networks using the following methods: 
(1) Standard: training with unperturbed data; 
(2) ADV: training with 10-step untargeted PGD examples 
(3) ADV-T: training with 10-step randomly targeted PGD examples
(4) TRADES and
(5) ADV-hCE: training with 10-step untargeted PGD examples based on a modified hierarchical cross-entropy (hCE) loss.

To further justify the use of separate networks for the coarse and fine classification tasks in HAR, we compare with two natural baselines.
ADV-T is a targeted-variant of the PGD adversarial training in \cite{madry2017towards}. 
Specifically, given a pair of input from the training set $(x,y)$ and $y \in z^*$, the perturbation is computed based on a targeted 10-step PGD attack where the target label is uniformly random sampled from $\{y \mid y \not \in z^*\}$.
%
%
An alternative to HAR is to train a single flat network with a modified hierarchical loss.
The idea of incorporating hierarchical structure in losses has been explored in the literature~\citep{cesa2006incremental, redmon2017yolo9000, ge2018deep, wehrmann2018hierarchical}, but not yet studied under the robust learning setting.
As such, we performed ADV-hCE, i.e., adversarial training using a hierarchical cross-entropy loss: $\ell_{\text{hCE}}(x, y, z) \triangleq \ell(F(x), y) + \ell(G(x), z)$ where $F(x)$ is the prediction of the network of the fine classes. $G(x)$ is the predictions of the coarse classes and its entries are computed by summing the corresponding fine class predictions in $F(x)$.

To differentiate HAR networks from the flat models, we refer the flat models as \emph{vanilla} models. 
We use a single ResNet50 for the vanilla model. 
For the proposed HAR network, we use multiple ResNet10 for the coarse network and the fine networks.
We use individual models with a lower capacity for HAR to reduce the difference in the order of magnitude of parameters between a single ResNet50 in the vanilla model and multiple ResNet10 in HAR. 
Notice that in the experiment with CIFAR-10 and CIFAR-100-5x5, the number of trainable parameters of the vanilla model is larger than that of the HAR network using the selected ResNet architectures.
While this is difficult to achieve with CIFAR-100, the goal here is to eliminate the concern of which the improved hierarchical adversarial robustness is obtained solely due to the increasing network complexity.
A comparison between the number of trainable parameters is included in \ifSupp Appendix~\ref{sec:App_parameter}\else the supplementary material \fi.
Note that in the HAR network, all component networks (CNN and FNNs) are trained using the same adversarial defence approach. 
As a concrete example, a HAR network trained with TRADES on CIFAR-100 consists of one coarse classifier and twenty fine classifiers where they are all trained using TRADES. 

For all five methods (Standard, ADV, ADV-T, ADV-hCE and TRADES), networks are trained for a total of 200 epochs, with an initial learning rate of 0.1. 
The learning rate decays by an order of magnitude at epoch 100 and 150. 
We used a minibatch size of 128 for testing and training. 
We used SGD optimizer with momentum of 0.9 and a weight decay of 2e-4. 
For TRADES, we performed a hyperparameter sweep on the strength of the regularization term $\beta$ and selected one that resulted in the highest accuracy against untargeted $\ell_\infty$ bounded PGD20 attacks. 
The optimization procedure is used for both the vanilla models and all component models in the HAR network.

\begin{table*}[t!]
\begin{footnotesize}
\caption{Accuracy of different models on CIFAR-100 against $\ell_{\infty}$ bounded white-box untargeted PGD attacks.}
\label{table:white_box_result}
\vskip 0.15in
\begin{center}
\setlength{\tabcolsep}{9pt} 
\renewcommand{\arraystretch}{1.3} 
\begin{tabular}{llrrrrrr}
\multicolumn{2}{c}{\multirow{2}{*}{Method}}                        & \multicolumn{2}{c}{Clean} & \multicolumn{2}{c}{PGD20 ($\eps=\frac{4}{255}$)} & \multicolumn{2}{c}{PGD20 ($\eps=\frac{8}{255}$)} \\ \cline{3-8}
\multicolumn{2}{c}{}                                               & Fine        & Coarse         & Fine        & Coarse         & Fine        & Coarse         \\ \Xhline{2\arrayrulewidth}
\multirow{4}{*}{Vanilla}                               & Standard  & $73.21\%$   & $82.57\%$      & $0.01\%$    & $24.89\%$      & $0.01\%$    & $24.89\%$      \\
                                                       & ADV       & $58.62\%$   & $69.81\%$      & $21.36\%$   & $37.80\%$      & $21.36\%$   & $37.80\%$      \\
                                                       & ADV-T     & $64.74\%$   & $75.02\%$      & $17.19\%$   & $41.30\%$      & $17.19\%$   & $41.30\%$      \\
                                                       & TRADES    & $57.12\%$   & $67.67\%$      & $26.69\%$   & $41.47\%$      & $26.69\%$   & $41.47\%$      \\ \hline \hline
\multirow{3}{*}{HAR}                                   & Standard  & $63.49\%$   & $81.24\%$      & $0.12\%$    & $29.25\%$      & $0.12\%$    & $29.25\%$      \\
                                                       & ADV       & $48.53\%$   & $66.23\%$      & $20.28\%$   & $30.53\%$      & $22.28\%$   & $30.53\%$      \\
                                                       & TRADES    & $46.62\%$   & $62.49\%$      & $22.00\%$   & $32.86\%$      & $22.00\%$   & $32.86\%$      \\\Xhline{5\arrayrulewidth}
\end{tabular}
\begin{tabular}{llrrrrrr}
\multicolumn{2}{c}{\multirow{2}{*}{Method}}                        & \multicolumn{2}{c}{PGD50 ($\eps=\frac{8}{255}$)} & \multicolumn{2}{c}{PGD100 ($\eps=\frac{8}{255}$)} & \multicolumn{2}{c}{PGD200 ($\eps=\frac{8}{255}$)} \\ \cline{3-8}
\multicolumn{2}{c}{}                                               & Fine        & Coarse         & Fine        & Coarse         & Fine        & Coarse         \\ \Xhline{2\arrayrulewidth}
\multirow{4}{*}{Vanilla}                               & Standard  & $0.01\%$    & $24.84\%$      & $0.00\%$    & $25.12\%$      & $0.01\%$    & $24.94\%$      \\
                                                       & ADV       & $21.05\%$   & $37.16\%$      & $20.87\%$   & $36.96\%$      & $20.94\%$   & $37.02\%$      \\
                                                       & ADV-T     & $16.94\%$   & $41.07\%$      & $16.88\%$   & $40.99\%$      & $16.79\%$   & $40.84\%$      \\
                                                       & TRADES    & $26.48\%$   & $41.22\%$      & $26.58\%$   & $41.40\%$      & $26.52\%$   & $41.27\%$      \\ \hline \hline
\multirow{3}{*}{HAR}                                   & Standard  & $0.15\%$    & $28.91\%$      & $0.14\%$    & $29.46\%$      & $0.15\%$    & $29.27\%$      \\
                                                       & ADV       & $19.91\%$   & $29.94\%$      & $20.03\%$   & $29.99\%$      & $19.89\%$   & $29.93\%$      \\
                                                       & TRADES    & $21.99\%$   & $32.66\%$      & $21.90\%$   & $32.60\%$      & $21.87\%$   & $32.36\%$      \\\Xhline{2\arrayrulewidth}
\end{tabular}
\end{center}
\end{footnotesize}
\end{table*}

\begin{table*}[t!]
\begin{footnotesize}
\caption{Accuracy of different models on CIFAR-100 against $\ell_{\infty}$ bounded worst-case targeted PGD attacks based on Algorithm \ref{alg:hierarchical_example}.}
\label{table:targeted_whitebox_linf}
\vskip 0.15in
\begin{center}
\setlength{\tabcolsep}{12pt} 
\renewcommand{\arraystretch}{1.3} 
\begin{tabular}{lrrrrrr}
\multicolumn{2}{c}{\multirow{2}{*}{Method}}                 & PGD20                  & PGD20                  & PGD50                 & PGD100                  & PGD200   \\ 
\multicolumn{2}{c}{}                                        & ($\eps=\frac{4}{255}$)   & ($\eps=\frac{8}{255}$)   & ($\eps=\frac{8}{255}$)  & ($\eps=\frac{8}{255}$)    & ($\eps=\frac{8}{255}$)  \\ \Xhline{2\arrayrulewidth}
\multirow{4}{*}{Vanilla}                     & Standard     & $0.00\%$               & $0.00\%$               & $0.00\%$              & $0.00\%$                & $0.00\%$ \\
                                             & ADV          & $43.30\%$              & $24.60\%$              & $24.60\%$             & $24.50\%$               & $24.00\%$ \\
                                             & ADV-T        & $43.50\%$              & $22.10\%$              & $21.70\%$             & $20.70\%$               & $21.00\%$ \\
                                             & TRADES       & $47.20\%$              & $30.00\%$              & $29.80\%$             & $29.70\%$               & $28.80\%$ \\ \hline \hline
\multirow{3}{*}{HAR}                         & Standard     & $8.60\%$               & $4.00\%$               & $3.50\%$              & $3.40\%$                & $3.30\%$ \\
                                             & ADV          & $\bm{43.70}\%$              & $\bm{25.80}\%$              & $\bm{25.50}\%$             & $\bm{25.30}\%$               & $\bm{24.40}\%$ \\
                                             & TRADES       & $45.80\%$              & $29.20\%$              & $28.90\%$             & $29.30\%$               & $\bm{28.90}\%$ \\ \Xhline{2\arrayrulewidth}

\end{tabular}
\end{center}
\end{footnotesize}
\end{table*}

\subsection{Hierarchical Robustness under Untargeted and Targeted Attacks}
\label{sec:exp_untargeted_targeted_attack}
There are several threat models to consider while evaluating adversarial robustness, regardless of standard or hierarchical robustness. 
The white-box threat model specifies that the model architecture and network parameters are fully transparent to the attacker ~\citep{goodfellow2014explaining}. 
Despite many white-box attack methods exist, perturbations generated using iterations of PGD remain as one of the most common benchmarks for evaluating adversarial robustness under the white-box setting. 
%
%
Specifically, we perform 20, 50, 100 and 200 iterations of PGD on the entire test set data for the untargeted attacks in Table \ref{table:white_box_result}.
%
%
To evaluate the hierarchical robustness of the model, we perform the worst-case hierarchical adversarial perturbations introduced in Section \ref{sec:generating_hierarchical_perturbation}.
Due to the large number of fine labels in CIFAR-100, the hierarchical attack was performed on 1000 randomly selected test set data.
%
%
%
All \LinfPGD{} adversarial examples used for all evaluations are generated a step size of $\eps/4$ (pixel values are normalized to $[0, 1]$).

Along with the two attacks, we also include results on unperturbed testset data (Clean). 
For clean and untargeted attacks, we report the percentage of correct fine class prediction as fine accuracy, and the percentage of fine class prediction belonging to the correct coarse class as coarse accuracy.
%
%
%
For the proposed targeted attack reported in Table \ref{table:targeted_whitebox_linf}, the accuracy refers to the percentage of the test set data where the targeted attack fails to alter the final prediction to the desired target, even after iterating through all eligible target labels. 
%
%
%
It is important to realize that a successful targeted attack implies misclassification for both coarse and fine classes. 
%
%

\subsubsection{Discussions}
Before making the comparison between the HAR model and the vanilla model, we make an interesting observation in Table \ref{table:white_box_result}: there is a consistent gap between the fine and coarse accuracy for both clean and perturbed inputs. This shows that under the untargeted attack, some of the misclassified inputs at the fine level are actually still classified correctly at the coarse level.
In particular, vanilla networks trained with unperturbed training data have close to $0\%$ fine accuracy under untargeted PGD attacks with $\eps = 8/255$, while nearly $25\%$ of all the misclassified inputs still belong to the correct coarse class. 
This shows that the untargeted attacks do not provide a good representation of the hierarchical adversarial robustness as an empirical evaluation, and it supports our earlier observation in Section \ref{sec:observations_untargeted_pgd} which is the main drive for the proposal of a targeted attack specifically designed for reducing coarse accuracy and damaging hierarchical robustness.

Indeed, results in Table \ref{table:targeted_whitebox_linf} show that the proposed worst-case targeted attack severely damages the hierarchical adversarial robustness of all models.
On Standard trained models, despite the high hierarchical robustness under untargeted attacks, nearly all of the CIFAR-100 data can be perturbed into a desired target class from another coarse class. 
Therefore, we emphasize the use of the proposed worst-case targeted attack for a more accurate empirical evaluations for the hierarchical adversarial robustness of the model.
For vanilla models trained with ADV and TRADES, we notice that the improved adversarial robustness on fine classes translates to an improvement in hierarchical adversarial robustness.

The result shows that ADV-hCE does not outperform the standard ADV in hierarchical robustness, and thus, we defer its discussions in \ifSupp Appendix \ref{sec:appendix_hCE}\else the supplementary material \fi.
We discuss ADV-T here, as it serves an excellent case to emphasize that since evaluating robustness empirically using attacks only shows an upper bound on the true robustness of the model, one should focus on the metric that results in the lowest upper bound. 
For this reason, a network outperforming another on weaker metrics does not imply it has better hierarchical robustness.
With this notion, we observed that vanilla models trained using ADV-T shows improved hierarchical robustness against untargeted PGD attacks compared to the original adversarial training method. 
%
%
However, results in Table \ref{table:targeted_whitebox_linf} suggests that the seeming improvement in hierarchical robustness of the ADV-T method does not hold against stronger hierarchical attacks.
ADV-trained models outperform ADV-T models on all targeted PGD attacks with $\eps = 8/255$.
%

Finally, the result shows that ADV-trained HAR network significantly improves the hierarchical robustness against all worst-case targeted attacks compared to the vanilla counterpart.
Particularly, we notice a $0.4\%$ increase in accuracy against PGD200 adversaries. 
For TRADES-robusitified networks, we observe that although the vanilla model outperforms HAR against weaker targeted adversaries (i.e., smaller $\eps$ and small PGD iterations), the gain in hierarchical robustness with HAR rises as the targeted attack becomes stronger.
For the strongest PGD200 attack, we notice a $0.1\%$ increase in accuracy compared to the vanilla model.

\subsection{Hierarchical Robustness under Targeted Attacks based on the Coarse Network}
\label{sec:exp_targeted_coarse}

Under the white-box threat model, attackers with a complete knowledge of the internal structure of HAR can also generate perturbations based on the coarse network. 
During evaluations, we investigate whether the targeted PGD adversaries based on the coarse network are stronger hierarchical adversarial examples compared to the ones generated using the entire network. 
%
%
Such attacks can be understood as finding a more general perturbation which alters the probability distribution of the coarse class: $P(z \mid x)$. 
Similar to the attack proposed in Section \ref{sec:generating_hierarchical_perturbation}, we perform an iterative, worst-case targeted PGD20 attack based on the coarse neural network. 
Specifically, we replace $\ell(F(x), y)$ with $\ell(G(x), z)$ in Equation \ref{eq:targeted_attack}, and iterate through all eligible coarse classes as target labels.
For example, to generate such attacks for HAR with ADV-trained component networks, the iterative targeted attack is performed based on the ADV-trained coarse network in the original HAR network.
%
%
%
Note that there is a distinction between the above attack procedure and a transfer-based attack where the perturbation is transferred from an independently trained source model \citep{papernot2017practical}. 
Since the perturbation is generated using part of the HAR network, such attacks still belongs in the white-box setting.
%
%
%
%
Results in Table~\ref{table:parent_attack} show that the perturbations generated using the coarse network are weaker attacks compared to the ones generated using the entire HAR network. 
%
%
%

\begin{table}[t!]
\begin{footnotesize}
\caption{Accuracy of the HAR network on CIFAR-100 against $\ell_{\infty}$ bounded targeted attacks ($\eps = 8/255$) generated using the coarse network (Coarse). For comparison, the untargeted attack (PGD200 from Table~\ref{table:white_box_result}) and the attack generated using the entire HAR network (PGD200 from Table~\ref{table:targeted_whitebox_linf}) are also included.}
\label{table:parent_attack}
\vskip 0.15in
\begin{center}
\setlength{\tabcolsep}{12pt} 
\renewcommand{\arraystretch}{1.3} 
\begin{tabular}{lllll}
Method    & Untargeted & Coarse        & HAR  \\ \Xhline{2\arrayrulewidth}
Standard  & $29.27\%$ & $\bm{0.00}\%$      & $3.30\%$  \\
ADV       & $29.93\%$ & $29.96\%$     & $\bm{24.40}\%$  \\
TRADES    & $32.36\%$ & $29.38\%$     & $\bm{28.90}\%$  \\\Xhline{2\arrayrulewidth}
\end{tabular}
\end{center}
\end{footnotesize}
\end{table}
\section{Conclusion}
\label{sec:Conclusion}
In this work, we introduced a novel concept called hierarchical adversarial examples.
For dataset which classes can be further categorized into fine and coarse classes, we defined hierarchical adversarial examples as the ones leading to a misclassification at the coarse level. 
We proposed a worst-case targeted PGD attack to generate hierarchical adversarial examples.
To improve the hierarchical adversarial robustness of DNNs, we proposed the HAR network design, a composite of a coarse network and fine networks where each component network is trained independently by adversarial defence techniques. 
We empirically showed that HAR leads to a significant increase in hierarchical adversarial robustness on CIFAR-10 and CIFAR-100.

The rapid adoption of machine learning applications has also led to an increasing importance in improving the robustness and reliability of such techniques. 
Mission-critical and safety-critical systems which rely on DNNs in their decision-making process shall incorporate robustness, along with accuracy, in the development process.
The introduction of the hierarchical adversarial examples and ways to defend against them is an important step towards a more safe and trustworthy AI system.

\bibliography{_reference}

\begin{thebibliography}{31}
\providecommand{\natexlab}[1]{#1}
\providecommand{\url}[1]{\texttt{#1}}
\expandafter\ifx\csname urlstyle\endcsname\relax
  \providecommand{\doi}[1]{doi: #1}\else
  \providecommand{\doi}{doi: \begingroup \urlstyle{rm}\Url}\fi

\bibitem[Carlini and Wagner(2017)]{carlini2017towards}
Nicholas Carlini and David Wagner.
\newblock Towards evaluating the robustness of neural networks.
\newblock In \emph{2017 ieee symposium on security and privacy (sp)}, pages
  39--57. IEEE, 2017.

\bibitem[Cesa-Bianchi et~al.(2006)Cesa-Bianchi, Gentile, and
  Zaniboni]{cesa2006incremental}
Nicolo Cesa-Bianchi, Claudio Gentile, and Luca Zaniboni.
\newblock Incremental algorithms for hierarchical classification.
\newblock \emph{The Journal of Machine Learning Research}, 7:\penalty0 31--54,
  2006.

\bibitem[Cordts et~al.(2016)Cordts, Omran, Ramos, Rehfeld, Enzweiler, Benenson,
  Franke, Roth, and Schiele]{cordts2016cityscapes}
Marius Cordts, Mohamed Omran, Sebastian Ramos, Timo Rehfeld, Markus Enzweiler,
  Rodrigo Benenson, Uwe Franke, Stefan Roth, and Bernt Schiele.
\newblock The cityscapes dataset for semantic urban scene understanding.
\newblock In \emph{Proceedings of the IEEE conference on computer vision and
  pattern recognition}, pages 3213--3223, 2016.

\bibitem[Croce and Hein(2020)]{croce2020reliable}
Francesco Croce and Matthias Hein.
\newblock Reliable evaluation of adversarial robustness with an ensemble of
  diverse parameter-free attacks.
\newblock In \emph{International Conference on Machine Learning}, pages
  2206--2216. PMLR, 2020.

\bibitem[Deng et~al.(2009)Deng, Dong, Socher, Li, Li, and
  Fei-Fei]{deng2009imagenet}
Jia Deng, Wei Dong, Richard Socher, Li-Jia Li, Kai Li, and Li~Fei-Fei.
\newblock Imagenet: A large-scale hierarchical image database.
\newblock In \emph{2009 IEEE conference on computer vision and pattern
  recognition}, pages 248--255. Ieee, 2009.

\bibitem[Deng et~al.(2011)Deng, Satheesh, Berg, and Li]{deng2011fast}
Jia Deng, Sanjeev Satheesh, Alexander~C Berg, and Fei Li.
\newblock Fast and balanced: Efficient label tree learning for large scale
  object recognition.
\newblock In \emph{Advances in Neural Information Processing Systems}, pages
  567--575, 2011.

\bibitem[Deng et~al.(2012)Deng, Krause, Berg, and Fei-Fei]{deng2012hedging}
Jia Deng, Jonathan Krause, Alexander~C Berg, and Li~Fei-Fei.
\newblock Hedging your bets: Optimizing accuracy-specificity trade-offs in
  large scale visual recognition.
\newblock In \emph{2012 IEEE Conference on Computer Vision and Pattern
  Recognition}, pages 3450--3457. IEEE, 2012.

\bibitem[Everingham et~al.(2015)Everingham, Eslami, Van~Gool, Williams, Winn,
  and Zisserman]{everingham2015pascal}
Mark Everingham, SM~Ali Eslami, Luc Van~Gool, Christopher~KI Williams, John
  Winn, and Andrew Zisserman.
\newblock The pascal visual object classes challenge: A retrospective.
\newblock \emph{International journal of computer vision}, 111\penalty0
  (1):\penalty0 98--136, 2015.

\bibitem[Ge(2018)]{ge2018deep}
Weifeng Ge.
\newblock Deep metric learning with hierarchical triplet loss.
\newblock In \emph{Proceedings of the European Conference on Computer Vision
  (ECCV)}, pages 269--285, 2018.

\bibitem[Geiger et~al.(2013)Geiger, Lenz, Stiller, and
  Urtasun]{geiger2013vision}
Andreas Geiger, Philip Lenz, Christoph Stiller, and Raquel Urtasun.
\newblock Vision meets robotics: The kitti dataset.
\newblock \emph{The International Journal of Robotics Research}, 32\penalty0
  (11):\penalty0 1231--1237, 2013.

\bibitem[Goodfellow et~al.(2015)Goodfellow, Shlens, and
  Szegedy]{goodfellow2014explaining}
Ian~J Goodfellow, Jonathon Shlens, and Christian Szegedy.
\newblock Explaining and harnessing adversarial examples.
\newblock In \emph{International Conference on Learning Representations}, 2015.

\bibitem[Gowal et~al.(2019)Gowal, Uesato, Qin, Huang, Mann, and
  Kohli]{gowal2019alternative}
Sven Gowal, Jonathan Uesato, Chongli Qin, Po-Sen Huang, Timothy Mann, and
  Pushmeet Kohli.
\newblock An alternative surrogate loss for pgd-based adversarial testing.
\newblock \emph{arXiv preprint arXiv:1910.09338}, 2019.

\bibitem[Hastie et~al.(2009)Hastie, Tibshirani, and
  Friedman]{HastieTibshiraniFriedman2009}
Trevor Hastie, Robert Tibshirani, and Jerome Friedman.
\newblock \emph{The Elements of Statistical Learning: Data Mining, Inference,
  and Prediction (2nd edition)}.
\newblock Springer, 2009.

\bibitem[He et~al.(2016)He, Zhang, Ren, and Sun]{he2016deep}
Kaiming He, Xiangyu Zhang, Shaoqing Ren, and Jian Sun.
\newblock Deep residual learning for image recognition.
\newblock In \emph{Proceedings of the IEEE conference on computer vision and
  pattern recognition}, pages 770--778, 2016.

\bibitem[Krizhevsky(2009)]{krizhevsky2009learning}
Alex Krizhevsky.
\newblock Learning multiple layers of features from tiny images.
\newblock 2009.

\bibitem[Kurakin et~al.(2017)Kurakin, Goodfellow, and
  Bengio]{kurakin2016adversarial}
Alexey Kurakin, Ian Goodfellow, and Samy Bengio.
\newblock Adversarial machine learning at scale.
\newblock In \emph{International Conference on Learning Representations}, 2017.

\bibitem[Li et~al.(2003)Li, Chang, and Wu]{li2003discovery}
Beitao Li, Edward Chang, and Yi~Wu.
\newblock Discovery of a perceptual distance function for measuring image
  similarity.
\newblock \emph{Multimedia systems}, 8\penalty0 (6):\penalty0 512--522, 2003.

\bibitem[Liu et~al.(2013)Liu, Sadeghi, Tappen, Shamir, and
  Liu]{liu2013probabilistic}
Baoyuan Liu, Fereshteh Sadeghi, Marshall Tappen, Ohad Shamir, and Ce~Liu.
\newblock Probabilistic label trees for efficient large scale image
  classification.
\newblock In \emph{Proceedings of the IEEE conference on computer vision and
  pattern recognition}, pages 843--850, 2013.

\bibitem[Madry et~al.(2018)Madry, Makelov, Schmidt, Tsipras, and
  Vladu]{madry2017towards}
Aleksander Madry, Aleksandar Makelov, Ludwig Schmidt, Dimitris Tsipras, and
  Adrian Vladu.
\newblock Towards deep learning models resistant to adversarial attacks.
\newblock In \emph{International Conference on Learning Representations}, 2018.

\bibitem[Mao et~al.(2021)Mao, Chen, Wang, Su, He, and Xue]{mao2020composite}
Xiaofeng Mao, Yuefeng Chen, Shuhui Wang, Hang Su, Yuan He, and Hui Xue.
\newblock Composite adversarial attacks.
\newblock In \emph{Proceedings of the AAAI Conference on Artificial
  Intelligence}, 2021.

\bibitem[Marsza{\l}ek and Schmid(2008)]{marszalek2008constructing}
Marcin Marsza{\l}ek and Cordelia Schmid.
\newblock Constructing category hierarchies for visual recognition.
\newblock In \emph{European conference on computer vision}, pages 479--491.
  Springer, 2008.

\bibitem[Papernot et~al.(2017)Papernot, McDaniel, Goodfellow, Jha, Celik, and
  Swami]{papernot2017practical}
Nicolas Papernot, Patrick McDaniel, Ian Goodfellow, Somesh Jha, Z~Berkay Celik,
  and Ananthram Swami.
\newblock Practical black-box attacks against machine learning.
\newblock In \emph{Proceedings of the 2017 ACM on Asia conference on computer
  and communications security}, pages 506--519, 2017.

\bibitem[Redmon and Farhadi(2017)]{redmon2017yolo9000}
Joseph Redmon and Ali Farhadi.
\newblock Yolo9000: better, faster, stronger.
\newblock In \emph{Proceedings of the IEEE conference on computer vision and
  pattern recognition}, pages 7263--7271, 2017.

\bibitem[Szegedy et~al.(2014)Szegedy, Zaremba, Sutskever, Bruna, Erhan,
  Goodfellow, and Fergus]{szegedy2013intriguing}
Christian Szegedy, Wojciech Zaremba, Ilya Sutskever, Joan Bruna, Dumitru Erhan,
  Ian Goodfellow, and Rob Fergus.
\newblock Intriguing properties of neural networks.
\newblock In \emph{International Conference on Learning Representations}, 2014.

\bibitem[Szegedy et~al.(2017)Szegedy, Ioffe, Vanhoucke, and
  Alemi]{szegedy2016inception}
Christian Szegedy, Sergey Ioffe, Vincent Vanhoucke, and Alexander Alemi.
\newblock Inception-v4, inception-resnet and the impact of residual connections
  on learning.
\newblock In \emph{Proceedings of the AAAI Conference on Artificial
  Intelligence}, volume~31, 2017.

\bibitem[Tashiro et~al.(2020)Tashiro, Song, and Ermon]{tashiro2020diversity}
Yusuke Tashiro, Yang Song, and Stefano Ermon.
\newblock Diversity can be transferred: Output diversification for white-and
  black-box attacks.
\newblock \emph{Advances in Neural Information Processing Systems}, 33, 2020.

\bibitem[Tousch et~al.(2012)Tousch, Herbin, and Audibert]{tousch2012semantic}
Anne-Marie Tousch, St{\'e}phane Herbin, and Jean-Yves Audibert.
\newblock Semantic hierarchies for image annotation: A survey.
\newblock \emph{Pattern Recognition}, 45\penalty0 (1):\penalty0 333--345, 2012.

\bibitem[Wehrmann et~al.(2018)Wehrmann, Cerri, and
  Barros]{wehrmann2018hierarchical}
Jonatas Wehrmann, Ricardo Cerri, and Rodrigo Barros.
\newblock Hierarchical multi-label classification networks.
\newblock In \emph{International Conference on Machine Learning}, pages
  5075--5084. PMLR, 2018.

\bibitem[Wong et~al.(2019{\natexlab{a}})Wong, Rice, and Kolter]{wong2020fast}
Eric Wong, Leslie Rice, and J~Zico Kolter.
\newblock Fast is better than free: Revisiting adversarial training.
\newblock In \emph{International Conference on Learning Representations},
  2019{\natexlab{a}}.

\bibitem[Wong et~al.(2019{\natexlab{b}})Wong, Schmidt, and
  Kolter]{wong2019wasserstein}
Eric Wong, Frank Schmidt, and Zico Kolter.
\newblock Wasserstein adversarial examples via projected sinkhorn iterations.
\newblock In \emph{International Conference on Machine Learning}, pages
  6808--6817. PMLR, 2019{\natexlab{b}}.

\bibitem[Zhang et~al.(2019)Zhang, Yu, Jiao, Xing, El~Ghaoui, and
  Jordan]{zhang2019theoretically}
Hongyang Zhang, Yaodong Yu, Jiantao Jiao, Eric Xing, Laurent El~Ghaoui, and
  Michael Jordan.
\newblock Theoretically principled trade-off between robustness and accuracy.
\newblock In \emph{International Conference on Machine Learning}, pages
  7472--7482. PMLR, 2019.

\end{thebibliography}
\bibliographystyle{plainnat}

\newpage
\appendix
\label{sec:Appendix}
\section{Comparison of trainable model parameters}
\label{sec:App_parameter}

In our evaluations with CIFAR-100, we use ResNet50 for the vanilla models and multiple ResNet10 for the HAR network. 
Note that we use architectures with a lower capacity for the individual components in HAR to reduce the difference in the order of magnitude of parameters between a single ResNet50 model and multiple ResNet10 models. 
For the evaluation with CIFAR-10, we use ResNet34 for the vanilla models, and use multiple ResNet10 for the HAR network. 
For the evaluation with CIFAR-100-5x5, we use ResNet50 for the vanilla models, and use multiple ResNet10 for the HAR network. 
In those two cases, the number of trainable parameters of the vanilla model is larger than that of the HAR network.
While this is difficult to achieve with CIFAR-100 given its large number of coarse labels, the goal here is to address the concern of which the improved hierarchical adversarial robustness is obtained only due to the increasing network complexity. 

\begin{table}[h]
\begin{footnotesize}
\caption{Number of trainable parameters in the ResNet architectures used in our evaluations}
\label{table:App_parameter}
\vskip 0.15in
\begin{center}
\setlength{\tabcolsep}{12pt} 
\renewcommand{\arraystretch}{1.3} 
\begin{tabular}{lr}
Model   	&  Number of Parameters 	  \\ \Xhline{2\arrayrulewidth}
ResNet10    & $4.9$ M   \\
ResNet34    & $21.3$ M   \\
ResNet50    & $25.6$ M   \\\Xhline{2\arrayrulewidth}
\end{tabular}
\end{center}
\end{footnotesize}
\end{table}

\ifArxiv\newpage\else\fi
\section{Hyperparameter sweep for TRADES on ResNet10}
\label{sec:App_hyperparameter_sweep}

An important hyperparameter in training with TRADES \citep{zhang2019theoretically} is $\beta$ which balances the trade-off between standard accuracy on the clean data and the robust accuracy on the perturbed one. 
In \citet{zhang2019theoretically}, the sweep of $\beta$ was only performed on more complex models such as WideResNet.
As such, we performed a similar hyperparameter sweep to determine the optimal $\beta$ in TRADES on ResNet10 using the CIFAR-10 dataset.
In Table~\ref{table:trades}, we compared the standard accuracy and the robust accuracy of models trained using TRADES with various value of $\beta$. 
The FGSM and the untargeted PGD20 perturbations are $\ell_{\infty}$ bounded with ($\eps = \frac{8}{255}$).
Based on the result, we used $\beta$ of value $9$ while training with TRADES in our evaluations as it resulted in the highest untargeted PGD20 accuracy.
%
%
%
We emphasize the main goal of the paper is to demonstrate the novel concept of hierarchical adversarial robustness including techniques to generate and defend against them, rather than reaching for SOTA performance.
Therefore, fine-tuning the optimization parameters including the $\beta$ for the CIFAR-100-5x5 and CIFAR-100 datasets would not be necessary and we simply used the same $\beta$ for the two datasets.
%

\begin{table}[h]
\begin{footnotesize}
\caption{Hyperparameter sweep of TRADES on ResNet10 using CIFAR-10. The FGSM and PGD20 perturbations are  $\ell_{\infty}$ bounded with ($\eps = \frac{8}{255}$). We used $\beta$ of value $9$ while training with TRADES in the evaluations as it resulted in the highest untargeted PGD20 accuracy. }
\label{table:trades}
\vskip 0.15in
\begin{center}
\setlength{\tabcolsep}{12pt} 
\renewcommand{\arraystretch}{1.3} 
\begin{tabular}{llllllllll}
$\beta$ & Clean     & FGSM      & PGD20                \\ \Xhline{2\arrayrulewidth}
15      & $72.80\%$ & $48.85\%$ & $44.80\%$   \\
13      & $73.46\%$ & $49.18\%$ & $45.09\%$   \\
11      & $74.46\%$ & $49.39\%$ & $44.86\%$   \\
9       & $75.61\%$ & $\bm{50.06}\%$ & $\bm{45.38}\%$  \\
7       & $76.74\%$ & $50.05\%$ & $45.04\%$   \\
5       & $78.39\%$ & $50.22\%$ & $44.40\%$   \\
3       & $80.20\%$ & $49.84\%$ & $42.53\%$   \\
1       & $83.69\%$ & $45.61\%$ & $35.27\%$   \\
0.5     & $84.00\%$ & $46.16\%$ & $34.40\%$   \\
0.1     & $\bm{84.64}\%$ & $41.12\%$ & $16.52\%$   \\ \Xhline{2\arrayrulewidth}
\end{tabular}
\end{center}
\end{footnotesize}
\end{table}

\newpage
\section{Hierarchical structure of classes within the CIFAR-10 dataset}
\label{sec:structure_cifar10}

To demonstrate the effectiveness of the proposed hierarchical attack in Section \ref{sec:generating_hierarchical_perturbation} and the improved hierarchical robustness with HAR, we performed an extensive empirical study on three datasets: the small-sized CIFAR-10 dataset with 2 coarse classes and 10 fine classes, the medium-sized CIFAR-100-5x5 dataset with 5 coarse classes and 25 fine classes, and the large-sized CIFAR-100 dataset with 20 coarse classes and 100 fine classes. The hierarchical structure of the datasets are shown in Table \ref{table:cifar10_cifar100}.

\begin{table*}[h]
\begin{footnotesize}
\caption{The hierarchical structure of classes within the CIFAR-10, CIFAR-100-5x5 and CIFAR-100 dataset}
\label{table:cifar10_cifar100}
\vskip 0.15in
\begin{center}
\ifArxiv
    \setlength{\tabcolsep}{6pt} 
    \renewcommand{\arraystretch}{1.2} 
\else
    \setlength{\tabcolsep}{15pt} 
    \renewcommand{\arraystretch}{1.3} 
\fi
\begin{tabular}{llll}
                            & Coarse labels      & Fine labels  \\ \Xhline{2\arrayrulewidth}
\multirow{2}{*}{CIFAR-10}    & Animals      & bird, cat, deer, dog, frog, horse     \\
                            & Vehicles     & airplane, automobile, ship, truck     \\ \hline \hline
\multirow{5}{*}{CIFAR-100-5x5}   & Fish         & aquarium fish, flatfish, ray, shark, trout     \\
                            & Vehicles     & bicycle, bus, motorcycle, pickup truck, train     \\
                            & People       & baby, boy, girl, man, woman     \\
                            & Trees        & maple, oak, palm, pine, willow     \\
                            & Insects      & bee, beetle, butterfly, caterpillar, cockroach \\ \hline \hline
\multirow{5}{*}{CIFAR-100}   & Aquatic mammals	& beaver, dolphin, otter, seal, whale \\
& Fish	& aquarium fish, flatfish, ray, shark, trout \\
& Flowers	& orchids, poppies, roses, sunflowers, tulips \\
& Food containers	& bottles, bowls, cans, cups, plates \\
& Fruit and vegetables	& apples, mushrooms, oranges, pears, sweet peppers \\
& Household electrical devices	& clock, computer keyboard, lamp, telephone, television \\
& Household furniture	& bed, chair, couch, table, wardrobe \\
& Insects	& bee, beetle, butterfly, caterpillar, cockroach \\
& Large carnivores	& bear, leopard, lion, tiger, wolf \\
& Large man-made outdoor things	& bridge, castle, house, road, skyscraper \\
& Large natural outdoor scenes	& cloud, forest, mountain, plain, sea \\
& Large omnivores and herbivores	& camel, cattle, chimpanzee, elephant, kangaroo \\
& Medium-sized mammals	& fox, porcupine, possum, raccoon, skunk \\
& Non-insect invertebrates	& crab, lobster, snail, spider, worm \\
& People	& baby, boy, girl, man, woman \\
& Reptiles	& crocodile, dinosaur, lizard, snake, turtle \\
& Small mammals	& hamster, mouse, rabbit, shrew, squirrel \\
& Trees	& maple, oak, palm, pine, willow \\
& Vehicles 1	& bicycle, bus, motorcycle, pickup truck, train \\
& Vehicles 2	& lawn-mower, rocket, streetcar, tank, tractor \\ \Xhline{2\arrayrulewidth}
\end{tabular}
\end{center}
\end{footnotesize}
\end{table*}

\newpage
\section{Results on CIFAR-100 with $\ell_2$ bounded attacks}
\label{sec:appendix_cifar100_l2}

In this section, we empirically study the effectiveness of the $\ell_2$ bounded hierarchical attack and the improved hierarchical robustness from HAR. 
The $\ell_2$ bounded worts-case targeted PGD attack is formulated in a similar fashion as the one illustrated in Algorithm \ref{alg:hierarchical_example} and the only difference is that the perturbation generated at each iteration of PGD is ensured to be within the $\ell_2$-ball of the original input.
Additionally, due to the large number of fine labels in CIFAR-100, the hierarchical attack was performed on 1000 randomly selected test set data.
Similar to the setup with $\ell_\infty$ attacks in Section \ref{sec:exp_untargeted_targeted_attack}, we compare various baselines where the model is robustified against $\ell_2$ attacks.
Note that we were not able to achieve reasonable robustness results with TRADES against $\ell_2$ attacks. 
One possible reason is that the hyperparameter sweep was based on the $\ell_{\infty}$ results, as such, the optimal value for $\beta$ is unsuitable in the $\ell_2$ setting. 
For this reason, we omit TRADES in this section. 

Similar to Section \ref{sec:exp_untargeted_targeted_attack}, we make two important observations.
First, we notice that the lowest coarse accuracy for the untargeted PGD attack ($\eps = 0.5$) is $53.59\%$ and $48.91\%$ respectively for the ADV-trained vanilla model and the HAR network. 
In Table \ref{table:targeted_whitebox_l2}, we observe that the proposed worst-case targeted PGD attack severely degrades the coarse accuracy for both models. In particular, for the vanilla model, the proposed attack results in a $10\%$ decrease in coarse accuracy.
This matches the result with $\ell_\infty$ attacks in Section \ref{sec:exp_untargeted_targeted_attack} and it shows that the proposed worst-case targeted PGD attack provides a more accurate empirical representation of the hierarchical adversarial robustness of the model.
Next, we observe that the targeted adversarial training does not improve the hierarchical robustness compared to the standard adversarial training, and ADV-trained HAR network improves the hierarchical adversarial robustness.
Particularly, we observe a $1.1\%$ improvement in the hierarchical robustness against $\ell_2$ bounded worst-case targeted attacks with the HAR design in comparison to the vanilla network. 
%


\begin{table*}[h]
\begin{footnotesize}
\caption{Accuracy of different models on CIFAR-100 against $\ell_{2}$ bounded white-box untargeted PGD attacks.  (a higher score indicates better performance)}
\label{table:cifar100_l2_results}
\vskip 0.15in
\begin{center}
\setlength{\tabcolsep}{9pt} 
\renewcommand{\arraystretch}{1.3} 
\begin{tabular}{llrrrrrr}
\multicolumn{2}{c}{\multirow{2}{*}{Method}}                        & \multicolumn{2}{c}{Clean} & \multicolumn{2}{c}{PGD20 ($\eps=0.25$)} & \multicolumn{2}{c}{PGD20 ($\eps=0.5$)} \\ \cline{3-8}
\multicolumn{2}{c}{}                                               & Fine        & Coarse         & Fine        & Coarse         & Fine        & Coarse         \\ \Xhline{2\arrayrulewidth}
\multirow{1}{*}{Vanilla}                               & Standard  & $73.21\%$   & $82.57\%$      & $3.05\%$    & $35.95\%$      & $0.29\%$    & $31.86\%$           \\
                                                       & ADV       & $64.38\%$   & $74.85\%$      & $50.04\%$   & $64.05\%$      & $36.35\%$   & $53.77\%$           \\
                                                       & ADV-T     & $68.73\%$   & $78.58\%$      & $48.78\%$   & $64.62\%$      & $31.41\%$   & $53.10\%$           \\\hline \hline
\multirow{1}{*}{HAR}                                   & ADV       & $56.96\%$   & $73.31\%$      & $43.87\%$   & $60.68\%$      & $31.91\%$   & $48.94\%$           \\ \Xhline{5\arrayrulewidth}
\end{tabular}
\begin{tabular}{llrrrrrr}
\multicolumn{2}{c}{\multirow{2}{*}{Method}}                        & \multicolumn{2}{c}{PGD50 ($\eps=0.5$)} & \multicolumn{2}{c}{PGD100 ($\eps=0.5$)} & \multicolumn{2}{c}{PGD200 ($\eps=0.5$)} \\ \cline{3-8}
\multicolumn{2}{c}{}                                               & Fine        & Coarse         & Fine        & Coarse         & Fine        & Coarse         \\ \Xhline{2\arrayrulewidth}
\multirow{3}{*}{Vanilla}                               & Standard  & $0.17\%$    & $32.17\%$      & $0.13\%$    & $32.55\%$      & $0.12\%$    & $32.07\%$      \\
                                                       & ADV       & $36.16\%$   & $53.64\%$      & $36.11\%$   & $53.61\%$      & $36.07\%$   & $53.59\%$      \\
                                                       & ADV-T     & $31.21\%$   & $52.97\%$      & $31.15\%$   & $52.82\%$      & $31.09\%$   & $52.78\%$      \\\hline \hline
\multirow{1}{*}{HAR}                                   & ADV       & $31.90\%$   & $48.78\%$      & $31.76\%$   & $48.70\%$      & $31.70\%$   & $48.91\%$      \\ \Xhline{2\arrayrulewidth}
\end{tabular}
\end{center}
\end{footnotesize}
\end{table*}

\begin{table*}[h]
\begin{footnotesize}
\caption{Accuracy of different models on CIFAR-100 against $\ell_{2}$ bounded worst-case targeted PGD attacks.  (a higher score indicates better performance)}
\label{table:targeted_whitebox_l2}
\vskip 0.15in
\begin{center}
\setlength{\tabcolsep}{12pt} 
\renewcommand{\arraystretch}{1.3} 
\begin{tabular}{lrrr}
\multicolumn{2}{c}{\multirow{2}{*}{Method}}                 & PGD20                  & PGD50                 \\ 
\multicolumn{2}{c}{}                                        & ($\eps=0.5$)           & ($\eps=0.5$)          \\ \Xhline{2\arrayrulewidth}
\multirow{3}{*}{Vanilla}                     & Standard     & $0.10\%$               & $0.10\%$              \\
                                             & ADV          & $43.60\%$              & $42.90\%$              \\
                                             & ADV-T        & $39.20\%$              & $38.50\%$              \\ \hline \hline
\multirow{1}{*}{HAR}                         & ADV          & $\bm{44.30}\%$         & $\bm{44.00}\%$        \\\Xhline{2\arrayrulewidth}
\end{tabular}
\end{center}
\end{footnotesize}
\end{table*}

\newpage
\section{Results on CIFAR-10 and CIFAR-100-5x5 with $\ell_\infty$ bounded attacks}
\label{sec:appendix_cifar10}

In this section, we examine the $\ell_\infty$ bounded hierarchical attack and the improved hierarchical robustness from HAR on the small-sized CIFAR-10 with 2 coarse classes and 10 fine classes, and the medium-sized CIFAR-100-5x5 with 5 coarse classes and 25 fine classes.
In Table \ref{table:white_box_result_cifar10}, we compare the HAR network to vanilla baselines using the accuracy on clean data, untargeted PGD20 adversaries and the targeted PGD hierarchical attack proposed in Section \ref{sec:generating_hierarchical_perturbation}.
We observe that HAR network trained with ADV and TRADES significantly improves the robustness against the worst-case targeted attacks compared to the vanilla counterparts. 
On CIFAR-10, HAR network achieves a $7.52\%$ and $1.47\%$ improvement when trained with ADV and TRADES respectively. 
On CIFAR-100-5x5, we observe a drastic improvement in hierarchical robustness using HAR: with more than $10\%$ increase in resistance to the worst-case targeted attacks compared to the vanilla models.

\begin{table*}[h]
\begin{footnotesize}
\caption{Performance on CIFAR-10 and CIFAR-100-5x5 against $\ell_{\infty}$ bounded attacks ($\eps = 8/255$): the untargeted PGD20 attacks (Untargeted) and the worst-case targeted hierarchical attacks (Targeted). (a higher score indicates better model robustness against specific attacks) }
\label{table:white_box_result_cifar10}
\vskip 0.15in
\begin{center}
\setlength{\tabcolsep}{9pt} 
\renewcommand{\arraystretch}{1.3} 
\begin{tabular}{llllllll}
                         & \multicolumn{2}{c}{\multirow{2}{*}{Method}} & \multicolumn{2}{c}{Clean} & \multicolumn{2}{c}{Untargeted} & \multirow{2}{*}{Targeted} \\ 
                         & \multicolumn{2}{c}{}                        					& Fine       & Coarse      & Fine        & Coarse       &                           \\ \Xhline{2\arrayrulewidth}
\multirow{6}{*}{\rotatebox{90}{CIFAR-10}} & \multirow{3}{*}{Vanilla}         & Standard  & $92.88\%$   & $98.93\%$     & $0.00\%$     & $\underline{79.73\%}$       & $0.00\%$                          \\
                         &                                 					& ADV       & $84.31\%$   & $97.49\%$      & $47.65\%$    & $87.40\%$       & $72.00\%$                          \\
                         &                                 					& TRADES    & $79.70\%$   & $95.69\%$      & $48.89\%$    & $88.15\%$       & $79.76\%$                          \\ \cline{2-8}
                         & \multirow{3}{*}{HAR}   							& Standard  & $92.19\%$   & $98.44\%$      & $0.00\%$     & $\underline{65.46\%}$       & $0.00\%$                           \\
                         &                                 					& ADV       & $79.80\%$   & $95.57\%$      & $43.82\%$    & $86.38\%$       & $\bm{79.52\%}$                          \\
                         &                                 					& TRADES    & $77.78\%$   & $94.49\%$      & $46.75\%$    & $86.07\%$       & $\bm{81.23\%}$                          \\  \hline \hline

\multirow{6}{*}{\rotatebox{90}{CIFAR-100-5x5}} & \multirow{3}{*}{Vanilla}        & Standard  & $72.80\%$   & $92.68\%$      & $0.00\%$     & $\underline{65.52\%}$       & $0.00\%$                          \\
                         &                                 					& ADV       & $61.96\%$   & $86.08\%$      & $24.12\%$    & $70.68\%$       & $42.20\%$                          \\
                         &                                 					& TRADES    & $59.28\%$   & $84.16\%$      & $28.36\%$    & $69.60\%$       & $48.04\%$                          \\ \cline{2-8}
                         & \multirow{3}{*}{HAR}   							& Standard  & $67.12\%$   & $94.36\%$      & $3.71\%$     & $\underline{56.72\%}$       & $0.11\%$                          \\
                         &                                 					& ADV       & $57.48\%$   & $89.16\%$      & $25.80\%$    & $63.32\%$       & $\bm{52.32\%}$                          \\
                         &                                 					& TRADES    & $55.44\%$   & $84.96\%$      & $32.80\%$    & $66.36\%$       & $\bm{58.96\%}$                          \\\Xhline{2\arrayrulewidth}
\end{tabular}
\end{center}
\end{footnotesize}
\end{table*}

\newpage
\section{Hierarchical cross-entropy loss}
\label{sec:appendix_hCE}

To further justify the use of separate networks for the robust hierarchical classification tasks in HAR, we compare two additional two baselines. 
In Section \ref{sec:exp_untargeted_targeted_attack}, we showed that the improvement in hierarchical robustness of the ADV-T trained vanilla models does not hold against stronger hierarchical attacks.
Another alternative to using separate networks is to train a single flat network with a modified hierarchical loss.
The idea of incorporating hierarchical structure in losses has been explored in the literature~\citep{cesa2006incremental, redmon2017yolo9000, ge2018deep, wehrmann2018hierarchical}, but not yet studied under the robust learning setting.
As such, we proposed ADV-hCE, i.e., adversarial training using a hierarchical cross-entropy loss. First, we define the hierarchical cross-entropy loss as
\begin{equation}
    \ell_{\text{hCE}}(x, y, z) \triangleq \ell(F(x), y) + \ell(G(x), z),
\end{equation}
where $F(x)$ and $G(x)$ are the predictions of the fine classes and coarse classes respectively.
In particular, $F(x)$ is the output of the neural network and entries in $G(x)$ are computed by summing the corresponding fine class predictions in $F(x)$.
For example, suppose we have four fine classes and two coarse classes and, given an input, the prediction of the network is $F(x) = [0.2, 0.4, 0.3, 0.1]$. If the first two entries in $F(x)$ correspond to the same coarse classes, then we have $G(x) = [0.6, 0.4]$. 
Intuitively, the second term in the hierarchical cross-entropy loss can be understood as a penalty for assigning probabilities of fine labels outside of the correct coarse class.
As such, we trained two flat ResNet10 models on the CIFAR-10 dataset using adversarial training based on the cross-entropy loss and the hierarchical cross-entropy loss respectively.
It is important to realize that since adversarial training involves replacing the clean training data with the perturbed data, this means the PGD adversaries used during ADV-hCE are generated based on the hierarchical cross-entropy loss.
%
%
We observe that although models trained using the modified loss shows improved hierarchical robustness against untargeted PGD attacks generated from both the original and the hierarchical cross-entropy loss. However, such an improvement in hierarchical robustness does not hold against stronger targeted hierarchical attacks, captured by the $50\%$ drop in accuracy against the worst-case targeted attack.

\begin{table*}[h]
\begin{footnotesize}
\caption{Accuracy of the model trained with the hierarchical cross-entropy loss on CIFAR-10 against $\ell_\infty$ bounded attacks ($\eps = 8/255$): the untargeted PGD20 attacks based on the cross-entropy loss (Untargeted-CE), the untargeted PGD20 attacks based on the hierarchical cross-entropy loss (Untargeted-hCE) and the proposed worst-case hierarchical attack (Targeted). }
\label{table:hCE}
\vskip 0.15in
\begin{center}
\setlength{\tabcolsep}{12pt} 
\renewcommand{\arraystretch}{1.3} 
\begin{tabular}{lllllllll}
\multirow{2}{*}{Method} & \multicolumn{2}{c}{Clean} & \multicolumn{2}{c}{Untargeted-CE} & \multicolumn{2}{c}{Untargeted-hCE}  & \multirow{2}{*}{Targeted}\\ 
& Fine       & Coarse       & Fine         & Coarse       & Fine        & Coarse         &       \\ \Xhline{2\arrayrulewidth}
ADV-CE       & $78.38\%$  & $91.57\%$    & $39.68\%$    & $84.11\%$    & $42.10\%$   & $77.97\%$    & $77.44\%$  \\
ADV-hCE      & $74.89\%$  & $95.56\%$    & $38.26\%$    & $87.4\%$    & $39.56\%$   & $81.12\%$    & $22.34\%$   \\\Xhline{2\arrayrulewidth}
\end{tabular}
\end{center}
\end{footnotesize}
\end{table*}

\end{document}